\documentclass[letterpaper, 10 pt, conference]{ieeeconf}  

\newcommand*{\rom}[1]{\expandafter\@slowromancap\romannumeral #1@}

\IEEEoverridecommandlockouts
\usepackage{graphics} 
\usepackage{epsfig} 
\usepackage{times}
\usepackage{amsmath} 
\usepackage{amssymb}
\usepackage{dirtytalk}
\usepackage{hyphenat}
\usepackage{romannum}
\usepackage{multirow}
\usepackage{multicol}
\usepackage{textcomp}
\usepackage{caption}
\usepackage{cite}
\usepackage{bm}

\usepackage[colorlinks = true, allcolors = black, citecolor=blue]{hyperref}

\title{\LARGE \bf
On the Human Control of a Multiple Quadcopters with a Cable-suspended Payload System
}

\author{Pratik Prajapati, Sagar Parekh, and Vineet Vashista
\thanks{Pratik Prajapati, Sagar Parekh, and Vineet Vashista$^*$ are with the Human-Centered Robotics Lab at IIT Gandhinagar, Gujarat, India as a PhD student, a project assistant, and an assistant professor respectively.
        {\tt\small vineet.vashista@iitgn.ac.in} }
       \thanks{$^*$corresponding author}
}
\begin{document}
\maketitle
\begin{abstract}
A quadcopter is an under-actuated system with only four control inputs for six degrees of freedom, and yet the human control of a quadcopter is simple enough to be learned with some practice. In this work, we consider the problem of human control of a multiple quadcopters system to transport a cable-suspended payload. The coupled dynamics of the system, due to the inherent physical constraints, is used to develop a leader-follower architecture where the leader quadcopter is controlled directly by a human operator and the followers are controlled with the proposed Payload Attitude Controller and Cable Attitude Controller. Experiments, where a human operator flew a two quadcopters system to transport a cable-suspended payload, were conducted to study the performance of proposed controller. The results demonstrated successful implementation of human control in these systems. This work presents the possibility of enabling manual control for on-the-go maneuvering of the quadcopter-payload system which motivates aerial transportation in the unknown environments.


\end{abstract}

\vspace{-0.2in}
\section{INTRODUCTION}

In recent years, the technological advancement in electronics has led to the development of small, light-weight, and cheap Unmanned Aerial Vehicles (UAVs) with quadcopter being the most popular configuration. Despite being an under-actuated system, the manual control of a quadcopter is quite simple and can be learned with some practice as seen in high speed First Person View (FPV) \cite{fpv} drone racing where a human operator performs extremely agile maneuvers with great expertise. With the addition of a cable-suspended payload, the manual control of a quadcopter is possible but may require a lot of practice for the operator to develop the coupled dynamics intuition. However, if the onboard controller can compensate the payload dynamics then human control of the system can be achieved with relative ease. This approach can further be extended to human control of a multiple quadcopters collaboratively transporting a cable-suspended payload system. Such a capability can be very useful for applications involving agriculture, warehouses, and impromptu payload transport tasks.



Various mechanisms have been used in the literature for the applications of carrying payload using quadcopters. Rigid links arms and grippers have been used in \cite{thomas2014toward, mellinger2011design, kim2013aerial, mellinger2013cooperative}. Some works demonstrated the case of directly mounting the payload to the quadcopter chassis \cite{loianno2017cooperative, nguyen2015aerial}. In other works \cite{m1,m2,TLee,gassner2017dynamic,masone2016cooperative,fink2011planning}, a cable-suspended payload is also used. One advantage of using cables is that the quadcopter's rotational dynamics remains unaffected and the system retains its agility. However, the cable-suspended payload swings during the flight, with the possibilities of large amplitudes, which can destabilize the entire system. Therefore, the control of a single quadcopter or multiple quadcopters transporting a cable-suspended payload need to account for the physical constraints due to the coupled dynamics.

For a quadcopter with a cable-suspended payload system, trajectory planning algorithms for generating feasible trajectories have been developed using different approaches in \cite{foehn2017fast,tangVijay}. In addition, to compensate for the payload oscillations, estimator-based control algorithms have been developed in \cite{ExternalWrench,Passivity}. Further, a system with an arbitrary number of quadcopters with a cable-suspended payload is shown to be differentially flat \cite{m1}, and the flatness property is used to generate dynamically feasible trajectories for the system. In \cite{wrenchControl}, a distributed wrench control method is utilized to allow autonomous transportation of payload using multiple quadcopters without explicit communication between peers. A geometric controller is developed in \cite{m2} to track the position of the payload while the quadcopters maintained a prescribed formation. Notably, these works focused on the autonomous tracking of a predetermined trajectory of the payload. However, the scope of the applications for a quadcopter can be further enhanced by facilitating an on-the-go manipulation of the desired motion by a human operator. This can possibly allow the system to navigate through unknown environments without the need for mapping.

For the collaborative control of a multiple quadcopters system, leader-follower control strategies have been proposed. In \cite{tagliabue2017robust,tagliabue2016collaborative}, an admittance control based strategy was implemented to enable leader-follower collaboration to transport a payload along a desired motion. Further, in [10], a vision based localization, where the follower quadcopter would rely on visual and inertial feedback, was used. 

In this paper, a system with \emph{n} number of quadcopters collaboratively transporting a cable-suspended payload is modeled using the Lagrangian mechanics on a manifold and variation based linearization is used to linearize the system. A control architecture where a human directly controls the leader quadcopter and two control schemes, namely Payload Attitude Controller (PAC), and Cable Attitude Controller (CAC), are developed to control the follower quadcopters. The goal of PAC is to minimize the payload oscillations and that of CAC is to minimize the oscillations of cables connected with the followers to enable stable flight of the system. The performance of the proposed control architecture is evaluated through experiments using two quadcopters with a cable-suspended payload and in simulation with three quadcopters.

\vspace{-0.1in}
\section{METHODS}\label{methods}
\subsection{Dynamical Model of the System}
\begin{figure}[!htb]
    \centering
    \includegraphics[scale=0.9]{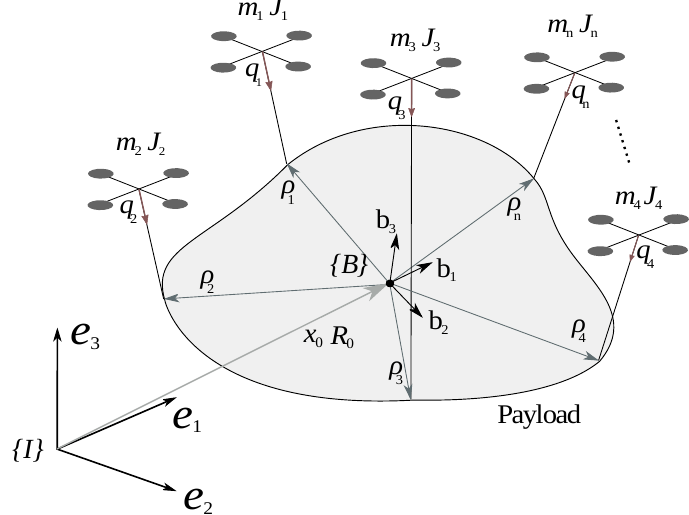}
    \caption{\small A system of multiple quadcopters collaboratively transporting a cable-suspended payload system. The fixed inertial frame is denoted by $\{I\}$ and the body-fixed frame of the payload is represented by $\{B\}$.\normalsize}
    \label{line_diagram_n_quadcopters}
\end{figure}
Consider a system of a cable-suspended payload being transported by multiple quadcopters collaboratively as shown in the Fig. \ref{line_diagram_n_quadcopters}. The body-fixed axes for the quadcopters are chosen as $\{b_{1_i}, b_{2_i}, b_{3_i} \}, i = 1,2,... n$ with the third axis being perpendicular to the plane of the quadcopter and pointing upwards. The position of the payload is defined as ${x_0} \in \mathbb{R} ^{3}$ and its orientation is denoted as a rotation matrix $R_0 \in SO(3)$ w.r.t frame $\{I\}$. ${\rho_{i}} \in \mathbb{R}^{3}, i= 1,2,.. n$ is the position vector from the centre of mass of the payload to the point of attachment of the cable of the $i^{th}$ quadcopter. $q_{i} \in \mathbb{S}^{2} , i = 1,2,... n$ represents the attitude of the $i^{th}$ cable and $R_i \in SO(3), i = 1,2,... n$ represents the attitude of the $i^{th}$ quadcopter. The configuration space of the system is given by $\mathbb{R}^{3} \times {SO(3)} \times (\mathbb{S}^{2} \times {SO(3)})^{n}$. $m_{i} \in \mathbb{R}$ and $J_{i} \in \mathbb{R}^{3 \times 3}$ denote the mass and the moment of inertia matrix of the $i^{th}$ quadcopter and $l_{i}$ is the length of the $i^{th}$ cable. The cables can be considered mass-less and if it remain taut then the position of the $i^{th}$ quadcopter in frame $\{I\}$ can be calculated as
${x_{i}} = {x_{0}} + R_{0} {\rho_{i}} - l_{i} q_{i}$.
The kinematics equations for each cable, quadcopter, and the payload are given as $\dot{q}_{i} = {\omega_{i}} \times q_{i}$, $\dot{R}_{i} = R_{i} {\widehat{\Omega}_{i}}$, and $\dot{R}_{0} =  R_{0} {\widehat{\Omega}_{0}}$ respectively. ${\omega_{i}} \in \mathbb{R}^{3}$ is the angular velocity of the $i^{th}$ cable. ${\Omega_{i}} \in \mathbb{R}^{3}$ is the angular velocity of the $i^{th}$ quadcopter in its body frame. ${\Omega_{0}} \in \mathbb{R}^{3}$ is the angular velocity of the payload in $\{B\}$. The $hat map \quad \widehat{\cdot}: \mathbb{R}^3 \xrightarrow{} so(3)$ is defined as ${\widehat{x}}{y} = {x} \times {y}$ for all ${x},{y} \in \mathbb{R}^3$ and $so(3)$ is the skew symmetric matrix \cite{goodarzi2016stabilization}. The total kinetic energy $(T)$ and the potential energy $(V)$ of the system are
\begin{equation}
\small
    T = \frac{1}{2} m_{0} {\Vert\dot{x}_{0} \Vert}^{2} + \frac{1}{2} \Omega_{0}^{T} J_{0} \Omega_{0}+ \frac{1}{2} \sum_{i=1} ^{n} m_{i} {\Vert \dot{x}_{i} \Vert}^{2} + \frac{1}{2} \sum_{i=1}^{n} \Omega_{i}^{T} J_{i} \Omega_{i}
    \label{kinetic_energy}
    \end{equation}
\begin{equation}
\small
    V = m_{0} g e_{3}^{T} x_{0}  + \sum _{i=1} ^{n} m_{i} g e_{3}^{T} x_{i} 
\label{potential_energy}
\end{equation}
\noindent
From the Lagrangian, $\mathcal{L}= T - V$, the infinitesimal variation\footnotemark
\, in the action integral can be written as
\noindent
\small
\begin{flalign}
\delta \mathcal{I} &= \int_{t_{0}}^{t_{f}} \delta \mathcal{L} \,\, dt
= \int_{t_{0}}^{t_{f}}\Bigg[  \frac{\partial \mathcal{L}}{\partial \dot{x}_{0}}^{T} \delta \dot{x}_{0} + \frac{\partial \mathcal{L}}{\partial x_{0}}^{T} \delta x_{0} + \frac{\partial \mathcal{L}}{\partial \Omega_{0}}^{T} \delta {\Omega}_{0} + \nonumber \\
& \frac{\partial \mathcal{L}}{\partial {R}_{0}}^{T} \delta {R}_{0} + \sum _{i=1}^{n} \Bigg( \frac{\partial \mathcal{L}}{\partial \dot{q}_{i}}^{T} \delta \dot{q}_{i} + \frac{\partial \mathcal{L}}{\partial {q}_{i}}^{T} \delta {q}_{i} + \frac{\partial \mathcal{L}}{\partial {\Omega}_{i}}^{T} \delta {\Omega}_{i} \Bigg)\Bigg] dt 
\end{flalign}
\normalsize
\footnotetext{The infinitesimal variation of $R\in SO(3)$, can be expressed in terms of the exponential map \cite{lee2017global} as
$\delta R = \frac{d}{d\epsilon}\Bigg|_{\epsilon=0}R \,\, exp(\epsilon\widehat{\eta}) \quad = \quad R\widehat{\eta} $, \,\,
where $\eta \in \mathbb{R}^3$. The infinitesimal variation in the body angular velocity, $\Omega \in \mathbb{R}^3$, and the cable attitude, $q \in S^2$, is given by
    $\delta \Omega = \widehat{\Omega} \eta + \dot{\eta} , \,\,\,\,
    \delta q = \frac{d}{d\epsilon} exp(\epsilon \widehat{\xi})  q=\widehat{\xi} \times q \nonumber$
where $\xi \in \mathbb{R}^3$ satisfies $\xi^{T} q = 0$. The corresponding infinitesimal variation in angular velocity of the cables is denoted as $\delta \omega$.}
\noindent
The total thrust force and moment generated by the $i^{th}$ quadcopter is denoted as $f_{i} \in \mathbb{R}$ and $M_{i} \in \mathbb{R}^{3}$ respectively. The components of this thrust force w.r.t. frame $\{I\}$ along $e_{1}$, $e_{2}$, and $e_{3}$ axes can be given as
\begin{equation}
    u_{i} =  f_{i} R_{i} e_{3} \in \mathbb{R}^{3},
\end{equation}
The variation in the virtual work \cite{goodarzi2016stabilization} with the variation in $i^{th}$ quadcopter position, $\delta x_{i}$, and quadcopter attitude, $\eta_{i}$ is given by
\begin{equation}
    \delta W=\int_{t_{0}}^{t_{f}} \sum_{i=1}^{n} \Big ( u_{i} \cdot \delta x_{i} + M_{i} \cdot \eta_{i} \Big) dt
\end{equation}
According to the Lagrange-d'Alembert Principle \cite{lee2017global},  the infinitesimal variation of the action integral over fixed time period is equal to negative of an infinitesimal variation of the work done by the external forces during the same time period, i.e. $\delta \mathcal{I} = - \delta W$. 

Using integration by part and rearranging the terms, the EOM of the system can be written as
\small
\begin{flalign}
2 M_{A} \ddot{x}_{0} + 2 \mathcal{M_{B}}( R_{0} {\widehat{\Omega}_{0}}^{2} \rho_{i} - R_{0} \widehat{\rho}_{i} \dot{\Omega}_{0} &-  l_{i} \ddot{q}_{i}) + M_{C} = \sum^{n}_{i=1} u_{i}\label{eom1} \\
J_{eq} \dot{\Omega}_{0} + 2 \mathcal{M_{B}} (\widehat{\rho}_{i} R_{0}^{T} \ddot{x}_{0} - \widehat{\rho}_{i} R_{0}^{T} l_{i} \ddot{q}_{i}) &= \nonumber \\ \sum_{i=1}^{n} \widehat{\rho}_{i}  R_{0}^{T} \big(  u_{i} - m_{i} g e_{3} \big) &- \widehat{\Omega}_{0} J_{eq} \Omega_{0} \label{eom2}\\
m_{i}l_{i}\ddot{q}_{i} + {\widehat{q}}^{2}_{i} m_{i} \ddot{x}_{0} - m_{i} {\ddot{q}}^{2}_{i} R_{0} \widehat{\rho}_{i} \dot{\Omega}_{0} &= \nonumber\\
{\widehat{q}}^{2}_{i} \big(  u_{i} - m_{i} R_{0} {\widehat{\Omega}}^{2}_{0} \rho_{i} - m_{i} g e_{3}  \big) &- m_{i} l_{i} {\Vert  \dot{q}_{i}  \Vert}^{2} q_{i}\\
J_{i} \dot{\Omega}_{i} + \Omega_{i} \times J_{i} \Omega_{i}
= M_{i} \label{eom5}
\end{flalign}
\normalsize
where $M_{A} = \frac{m_{0}}{2} + \frac{1}{2} \sum ^{n}_{i=1} m_{i}, \,\, \mathcal{M_{B}} = \frac{1}{2} \sum^{n}_{i=1} m_{i}$, and $M_{C} = m_{0} g e_{3} + \sum_{i=1} ^{n} m_{i} g e_{3}$


\subsection{Control Architecture}
\label{control_design_section}


Consider the problem of transporting a payload using a system of multiple quadcopters. As a human cannot control multiple quadcopters at a time, a leader-follower strategy is adopted in which human controls one of the quadcopters (leader) and the other quadcopters (followers) are autonomously controlled to facilitate a stable flight. As the payload is suspended from the quadcopters using cables, the movements of the leader quadcopter induces oscillations in the cables and the payload. The task of the follower quadcopters is to reduce these oscillations and maintain the payload in a desired orientation, to enable human control of the multiple quadcopters system to transport the payload. To achieve this, a control architecture is developed consisting of (1) Payload Attitude Controller (PAC) which reduces the payload oscillations, and (2) Cable Attitude Controller (CAC) which reduces the cable oscillations.



 


From Eqs. (\ref{eom1}-\ref{eom5}), the thrust force ($f_{i}$) generated by $i^{th}$ quadcopter is only along its third body-fixed axis, i.e. $b_{3_i}$, implying that the translational dynamics are under-actuated. Initially, for defining the control architecture, a fictitious thrust input, $u_i\in \mathbb{R}^{3}$, have been considered without the rotational dynamics.

At the hover equilibrium configuration, the attitude of the payload, cables, and quadcopters are $R_{0_e} = I_{3 \times 3}$, $q_{i_e}=-e_{3}$, and $R_{i_e} = I_{3 \times 3}$ respectively w.r.t frame $\{I\}$, and the position of the payload can be chosen arbitrarily, $x_{0_e}$. Further, the translational and angular velocities of the payload are zero, i.e. $\dot{x}_{0_e} = 0_{3 \times 1}$ and ${\Omega_0}_{e} = 0_{3 \times 1}$, and the angular velocities of the cables and quadcopters are zero, i.e. $\omega_{i_e}= 0_{3 \times 1}$ and $\Omega_{i_e} = 0_{3 \times 1}$. The position of the $i^{th}$ quadcopter at hover equilibrium configuration can be written as $x_{i_e} = x_{0_e} + \rho_{i} - l_{i} e_{3}$. The fictitious thrust input and control moment at equilibrium configuration with $f_{i_e} = (m_{i} + \frac{m_0}{n})g$ will be
\begin{equation}
    u_{i_e} = f_{i_e} R_{i_e} e_{3}, \,\,\, M_{i_e} = 0_{3 \times 1}
    \label{delta_u_i_e}
\end{equation}
As the dynamics of the system evolve on a nonlinear manifold, variation based linearization \cite{wu2015variation} is used to linearize the system dynamics about the hover equilibrium configuration. Consider the trajectory of the payload is denoted as $x_{0_d}$, which is not defined apriori and depends on the human. The variation in the payload position is given as $\delta x_0 = x_{0} - x_{0_d}$. Similarly, the variations in the other states of the system at hover equilibrium configuration are given as
\small
\begin{eqnarray}
Cables:&  \delta {q}_{i} = \xi_{i} \times e_3, \quad & \delta \dot{q}_{i} = \dot{\xi}_{i} \times e_3 \\
Payload:&  \delta R_{0} = R_{0_e} \widehat{\eta}_0 = \widehat{\eta}_0,  \quad & \delta \Omega_{0} = \dot{\eta}_{0}  \\
Quadcopters:&  \delta R_{i} = R_{i_e} \widehat{\eta}_i = \widehat{\eta}_i,  \quad & \delta \Omega_{i} = \dot{\eta}_{i}
\end{eqnarray}
\normalsize
\noindent
where $\eta_0 \in \mathbb{R}^{3}$ and $\xi_i \in \mathbb{R}^3$ which satisfies $\xi_i\cdotp e_3 = 0$. The variation of the angular velocity of the $i^{th}$ cable, $\omega_i$, is $\delta\omega_i \in \mathbb{R}^3$ such that $\delta\omega_i\cdotp e_3 = 0$. So, the third elements of $\xi_{i}$ and $\delta \omega_{i}$ are zero for hover equilibrium configuration. Hence, the state vector in the linearized model of the system contains $C^{T} \xi_i \in \mathbb{R}^{2}$, where $C = [ e_{1}, e_{2} ] \in \mathbb{R}^{3 \times 2}$. With the variation in the control input, $\delta u_{i} = u_{i} - u_{i_{e}}$, the linearized equations of motion are written as
\begin{equation}
    M\Ddot{x} + Gx = B\delta u + g(x, \dot{x})\label{linearized_model}
\end{equation}
Expressions for matrices $M,G$, and $B$ are omitted due to page constraint and made available on request. $g(x, \dot{x})$ represents the higher order terms which equal to zero for the chosen equilibrium configuration. The state vector, $x=[\delta x_0, \eta_{0},C^{T}\xi_{1},C^{T} \xi_{2},\cdots,C^{T} \xi_{n}]^{T}\in\mathbb{R}^{6+2n}$ and $\delta u = [\delta u_{1}, \delta u_{2},\cdots,\delta u_{n}]^{T} \in \mathbb{R}^{3n}$. The linearized dynamics of the system in state space is written as
\begin{equation}
    \dot{z} = A_0 z + B_0 \delta u 
    \label{state_space_form}
\end{equation}
where $z = [ x , \dot{x} ]^{T} \in \mathbb{R}^{2(6+2n)}$, $A_0 \in \mathbb{R}^{2(6+2n)\times2(6+2n)}$, and $B_0 \in \mathbb{R}^{2(6+2n)\times(3n)}$ are given by
\begin{eqnarray}
\small
    A_0 = \begin{bmatrix} \nonumber
            0_{(6+2n)\times (6+2n)} & I_{ (6+2n) \times (6+2n)} \\
            -M^{-1}G & 0_ { (6+2n) \times (6+2n) } 
          \end{bmatrix} , \nonumber
    B_0 = \begin{bmatrix}
            0_{(6+2n)\times (3n)} \\
            M^{-1}B
          \end{bmatrix} \nonumber 
\end{eqnarray}
\normalsize
\noindent


Consider the first quadcopter as the leader, i.e. fully controlled by a human. The value of $\delta u_{1}$ in Eq. (\ref{state_space_form}) is governed by the human and can be thought of as based on system's states with a feedback gain, $k_{h}$, i.e.
\begin{equation}
    \delta u_{1} = - k_h z   
    \label{u_1_human}
\end{equation}
For calculating the control inputs for the follower quadcopters, $\delta u_{i} \,\,\, \forall \,\, i \in [2, n]$, a state feedback controller is designed as
\begin{flalign}
\delta u_{i} =\delta u_{i_{,PAC}} + \delta u_{i_{,CAC}}
\label{state_feed_back_eq}
\end{flalign}
$\delta u_{i_{,PAC}}$ defines the PAC input, which reduces the payload oscillations, and $\delta u_{i_{,CAC}}$ defines the CAC input which reduces the cable oscillations.
\begin{flalign}
    \delta u_{i_{,PAC}} &= -(K_{\eta_{0}} \eta_{0} + K_{\dot{\eta}_{0}} \dot{\eta}_{0}) :=-K_{i_{,PAC}} z \label{pac_eq}\\
    \delta u_{i_{,CAC}} &= -(K_{\xi_{i}} C^{T} \xi_{i} + K_{\dot{\xi}_{i}} C^{T} \dot{\xi}_{i}) :=-K_{i_{,CAC}} z \label{cac_eq}
\end{flalign}
where $K_{\eta_0}, K_{\dot{\eta}_{0}} \in \mathbb{R}^{3\times3}$, \,\,$ K_{\xi_{i}},K_{\dot{\xi}_{i}} \in \mathbb{R}^{3\times2}$, and $K_{i_{,PAC}},K_{i_{,CAC}} \in \mathbb{R}^{3\times2(6+2n)}$ are the gains matrices. From Eq. (\ref{state_space_form}),
\begin{equation}
   \dot{z} = (A_0 - B_0 K) z
\end{equation}
where $K = \begin{bmatrix} k_h\,\, (k_{2_{,PAC}} + k_{2_{,CAC}})\,\, ... \,\, (k_{n_{,PAC}} + k_{n_{,CAC}}) \end{bmatrix}^{T}$

For $(A_0 - B_0 K) $ hurwitz, the equilibrium configuration will be asymptotically stable. This can be achieved by appropriately selecting PAC and CAC gains for non-aggressive maneuvers by human.

As the attitude dynamics of the payload evolves on $SO(3)$ \cite{lee2017global}, the actual errors between the desired attitude, $R_{0_d}$, and current attitude, $R_{0}$, and the error in the angular velocity can be defined as $e_{R_0} = \frac{1}{2} (R^{T}_{0_d} R_0 - R_0^{T} R_{0_d})^{\vee}, \,\, e_{\Omega_0} = \Omega_0 - (R_0^{T} R_{0_d})\Omega_{0_d}$. $\Omega_0 = R_0^{T} \dot{R_0} $ and $\Omega_{0_d} = R^{T}_{0_d} \dot{R}_{0_d}$, and  ${vee map\,\, (\cdot)^{\vee} : so(3) \xrightarrow{} \mathbb{R}^{3}}$.

For payload orientation, $R_0$, close to the desired orientation, $R_{0_d}$, the state [$\eta_0$, $\delta\Omega_0]^{T}$ can be approximated as the error between the desired and actual states of the system evolving on $SO(3)$ \cite{wu2015variation}.
\begin{eqnarray}
    \begin{bmatrix}
        \eta_0 & \delta \Omega_0
    \end{bmatrix} ^{T}
    \approx
    \begin{bmatrix}
        e_{R_0} & e_{\Omega_0}\end{bmatrix} ^{T}
    \label{errorinso3}
\end{eqnarray}


The attitude dynamics of the cables evolves on $\mathbb{S}^{2}$ \cite{lee2017global}. The actual error in the $i^{th}$ cable attitude, $e_{q_i}$, between the current attitude, $q_i$, and the desired attitude, $q_{d_i}$, and the error in the angular velocity can be written as $e_{q_i}=\widehat{q}_{d_i} q_i$, and $e_{\omega_i} =\omega_i - (- \widehat{q}_i^{2}) \omega_{d_i}$ respectively. For the actual attitude, $q_i$, of the $i^{th}$ cable close to the desired attitude, $q_{d_i}$, the state $[ \xi_i  , \,\delta \omega_i ]^{T}$ can be approximated as the error between the desired and actual states of the system evolving on $\mathbb{S}^{2}$. 
\begin{eqnarray}
        \begin{bmatrix}
        \xi_i &
        \delta \omega_i
    \end{bmatrix} ^{T}
    \approx
    \begin{bmatrix}
        e_{q_i} &
        e_{\omega_i}
    \end{bmatrix} ^{T}
    \label{errorins2}
\end{eqnarray}
Eqs. (\ref{errorinso3}) and (\ref{errorins2}) are used for defining the errors for developing the control input as given in Eq. (\ref{state_feed_back_eq}). The desired thrust vector, $u_{i_d} \in \mathbb{R}^3$, for the $i^{th}$ quadcopter can be written in terms of control input, $\delta u_i$, and fictitious thrust input at equilibrium, $u_{i_e}$ from Eq. (\ref{delta_u_i_e}), as
\begin{equation}
    u_{i_d} = u_{i_{e}} + \delta u_{i} \label{force_1}
\end{equation}


Thus, to enable the human control of the entire system, follower quadcopters needs to generate desired thrust, $u_{i_d} \in \mathbb{R}^{3} \,\,\, \forall \,\, i \in [2, n]$ in frame $\{I\}$, but a quadcopter can generate a thrust force, $f_{i} \in \mathbb{R}$, only along the third body fixed axis, i.e. $b_{3_i}$. However, utilizing the fact that the rotational dynamics of a quadcotper is fully actuated, $M_{i} \in \mathbb{R}^{3}$ from Eq. (\ref{eom5}), the quadcopter can be oriented at desired attitude such that the thrust force, $f_{i}$, along the axes of the frame $\{I\}$ provides the desired thrust, $u_{i_d}$. In this work, this is achieved by the roll and pitch motion of the quadcopter. 

The linearized model of the translational dynamics of a quadcopter \cite{mellinger2012trajectory}, keeping the initial yaw angle zero, i.e. $\psi_i= 0$, is given as
\begin{equation}
    m_{i} \ddot{x}_i= \begin{bmatrix}
    m_{i} g\theta_i \\
    -m_{i} g\phi_i \\
     f_i \end{bmatrix} \label{force_2}
\end{equation}
where $\phi_i$ and $\theta_i$ are the roll and pitch angles of the $i^{th}$ quadcopter respectively, and $f_i$ is the trust force generated by the $i^{th}$ quadcopter. From Eq. (\ref{force_1}) and Eq. (\ref{force_2}), the desired roll angle, $\phi_{d_i}$, pitch angle, $\theta_{d_i}$, and thrust force, $f_{i}$, can be calculated as
\begin{equation}
    \begin{bmatrix}
        \theta_{d_i} \\
        \phi_{d_i} \\
        f_{i}
    \end{bmatrix} =
    \begin{bmatrix}
        u_{i_{d_1}} / m_i g \\
        u_{i_{d_2}} / m_i g \\
        u_{i_{d_3}}
    \end{bmatrix}, \,\,\, \forall \,\, i \in [2, n] \label{des_inputs}
\end{equation}
For the $i^{th}$ follower quadcopter, a PID Attitude Controller is used to calculate the roll, pitch, and yaw moments, $M_{1_i}$, $M_{2_i}$, and $M_{3_i}$ respectively, to achieve the desired attitude.
\footnotesize
\begin{flalign}
\begin{bmatrix}
M_{1_i} \\ M_{2_i} \\ M_{3_i}
\end{bmatrix} = 
\begin{bmatrix}
k_{p_{\phi}} (\phi_{d_i} - \phi_{i}) + k_{i_{\phi}} \int (\phi_{d_i} - \phi_{i})dt+ k_{d_{\phi}} (\dot{\phi}_{d_i} - \dot{\phi}_{i}) \\ k_{p_{\theta}} (\theta_{d_i} - \theta_{i})+ k_{i_{\theta}} \int (\theta_{d_i} - \theta_{i}) dt + k_{d_{\theta}} (\dot{\theta}_{d_i} - \dot{\theta}_{i})  \label{cc2} \\
k_{p_{\psi}} (\psi_{d_i} - \psi_{i})+ k_{i_{\psi}} \int (\psi_{d_i} - \psi_{i}) dt + k_{d_{\psi}} (\dot{\psi}_{d_i} - \dot{\psi}_{i}) 
\end{bmatrix}
\end{flalign}
\normalsize

For evaluating the performance of the proposed controller, configuration error functions for the cables and the payload are defined on the manifolds $\mathbb{S}^2$ and $SO(3)$ respectively, adopted from \cite{bullo}. These configuration error functions are given as
\begin{equation}
    \begin{aligned}
         \text{For Cables} && \Psi_{q_i} &= 1 - q^{T}_{d_{i}} q_{i}, && i \in [1, n] \\
         \text{For Payload} && \Psi_{R_0} &= \frac{1}{2}tr(I - R_d^TR)
    \end{aligned}
    \label{errorFN}
\end{equation}

\section{EXPERIMENTAL SETUP}

\begin{figure}[!htb]
    \centering
    \includegraphics[scale=0.9]{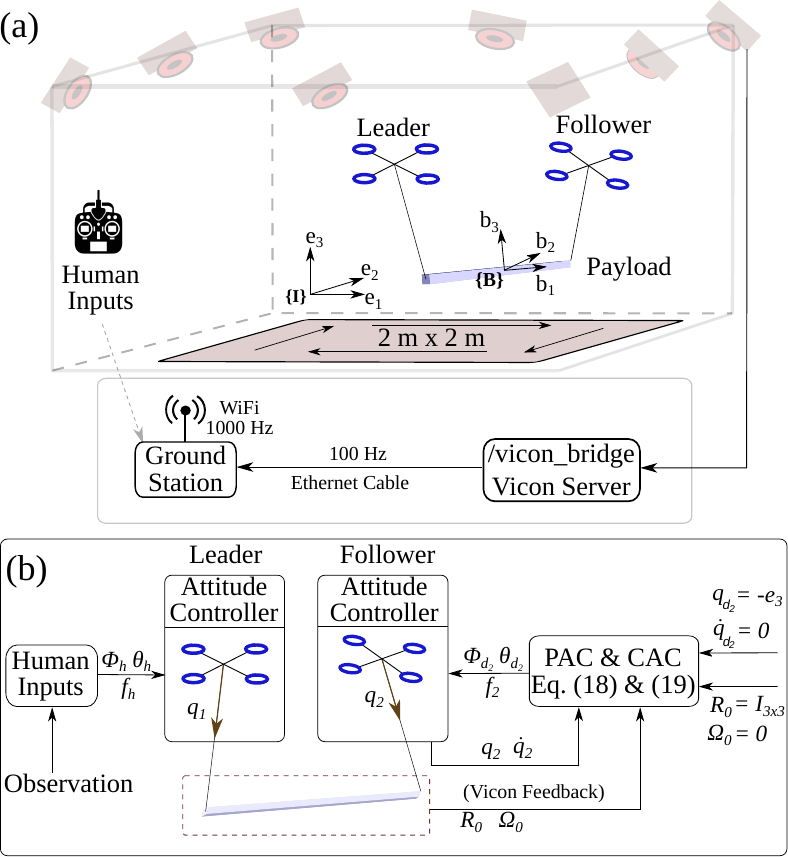}
    \caption{\small The communication and control architecture of the two quadcopters with a cable-suspended payload system are shown in (a) and (b) respectively. The leader quadcopter is fully controlled by the human with $\phi_h$, $\theta_h$, and $f_h$ as the human inputs. Based on the feedback of the attitude and the angular velocities of cable, ($q_{2}, \dot{q}_{2}$), and that of the payload, ($R_{0}, \Omega_{0}$), the desired roll angle, $\phi_d$, pitch angle, $\theta_d$, and thrust $f$, are calculated. \normalsize}
    \label{control_structure_two}
\end{figure}

The experimental setup used to implement the proposed control architecture is shown in Fig. \ref{control_structure_two}. PlutoX drones by Drona Aviation Pvt Ltd \cite{plutodrone}, frame size of $13 \, cm$, weight $52 \, g$, maximum payload carrying capacity of $15 \, g$, were used for conducting the experiments. Vicon\textsuperscript{\textregistered} motion capture system was used to record the real time flight data, which are used to estimate the attitude and angular velocity of the cables and the payload. Robot Operating System (ROS) was used for developing the control algorithm. The data from the Vicon system were communicated to the ground station through Ethernet cables using the ROS /vicon\_bridge package \cite{Vicon_bridge} at a rate of 100 Hz. The quadcopters were operated with the PID Attitude Controller which receives the control inputs from the ground station in terms of roll angle, pitch angle, yawing rate, and throttle via WiFi at a frequency of 1000 Hz. Yaw inputs were not given to the quadcopter in the presented experiments.

An experimental task was designed to conduct human operated quadcopter experiments where two quadcopters were flown using the leader-follower strategy using the proposed control architecture. Quadcopter 1, leader, is fully controlled by a human and quadcotper 2, follower, is autonomously controlled using CAC and PAC. A $60 \, cm$ long MDF strip, with cross-section  $1.5 \, cm \times 0.3 \, cm$ and mass of $24 \, g$, was used as the payload. Each end of the payload was suspended from a point slightly below from the CG of the respective quadcopters with $50 \, cm$ long cables. The first body fixed axis, $b_{1}$, of the payload is taken along the length and the third body fixed axis, $b_{3}$, is taken perpendicular to the length and along the thickness of the payload. The communication and control architecture for the system is shown in Fig. \ref{control_structure_two} (a) and (b) respectively. The task involved commanding the leader quadcopter such that the payload moves along a square trajectory within an area of $2 \, m \times 2 \, m$ in space. The human inputs are in terms of $\phi_{h}$, $\theta_{h}$, and $f_{h}$ which correspond to $\delta u_{1}$ in Eq. (\ref{u_1_human}). For the follower quadcopter, the control input, $\delta u_2$, is computed from PAC and CAC as per Eqs. (\ref{pac_eq}-\ref{cac_eq}) and is used to compute the desired roll angle, $\phi_{d_2}$, pitch angle, $\theta_{d_2}$, and thrust, $f_2$, using Eq. (\ref{des_inputs}). These desired values for both quadcopters serve as inputs to the onboard PID Attitude Controller of each quadcopter.

\section{RESULTS AND DISCUSSION}
\subsection{Human Experiments}
\vspace{-0.1in}
\begin{figure}[!htb]
    \centering
    \includegraphics[scale=0.95]{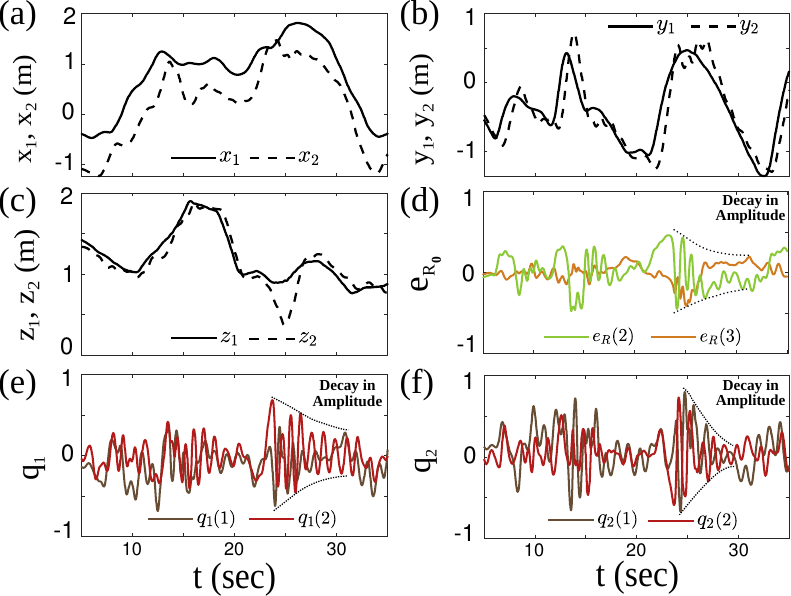}
    \caption{\small Experimental results. Plots (a-c) compare the positions of the leader,  $x_1$, $y_1$, and $z_1$, and follower, $x_2$, $y_2$, and $z_2$, along $e_1$, $e_2$, and $e_3$ axes respectively. The plot in (d) shows the variation in the error of the payload attitude, $e_{R_0}$, about the $b_2$ and $b_3$ axes. Plots (e) and (f) show the variation in the cable attitude, $q_i$, for the leader and follower along the $e_1$ and $e_2$ axes respectively \normalsize}
    \label{results_multiple}
\end{figure}
First, the experiment of two quadcopter-payload system was carried out using only onboard attitude controller without implementing CAC and PAC. The motion of the leader quadcopter induced oscillations in the cables and the payload. Without additional controller, the follower quadcopter did not receive the desired roll angle, $\phi_{d_2}$, pitch angle, $\theta_{d_2}$, and thrust force, $f_2$, to mitigate these oscillations. Hence, it was observed that for initial few seconds the follower quadcopter hovered after which the entire system crashed. Thus, the manual flight of two quadcopters-payload system is not feasible without additional control scheme.

The experimental results of a $30 \, s$ period of the human controlled flight are shown in Fig. \ref{results_multiple}. Figure \ref{results_multiple} (a-c) make a comparison between the position of the leader and follower quadcopters along $e_1, e_2$, and $e_3$ axes. The position of the follower quadcopter along $e_1$, i.e. $x_2$, is observed to be at an almost consistent offset of $60 \, cm$ from the leader quadcopter, i.e. $x_1$, which is the length of the payload. The position of the follower quadcopter along $e_2$ axis, i.e. $y_2$, and $e_3$ axis, i.e. $z_2$, coincides with that of the leader's except for the occasional overshoots when there is a sudden change in direction of motion. This implies that the proposed control architecture keeps the follower quadcopter behind the leader separated by the payload's length along $e_1$ and at the same level as the leader in the $e_2$-$e_3$ plane to keep the payload in the desired orientation, $R_{0_d} = I_{3\times3}$.

Figure \ref{results_multiple} (d) shows the error in the payload attitude, $e_{R_0}$, about $b_2$ and $b_3$ axes. As the cable attachments points on the payload form the $b_1$ axis, the error in the payload attitude about $b_1$ need not be controlled. As the leader quadcopter is continuously moving, it causes oscillations in the payload attitude, these oscillations are dampened by PAC as it maintains the payload in its desired orientation. The plots in Fig. \ref{results_multiple} (e) and (f) show the cable attitude for the leader and the follower quadcopters. It is observed that there is a rapid decay in the amplitude of the cable oscillations for the follower quadcopter, $q_2$. Notably, cable oscillations of leader quadcopter, $q_1$, are also observed to decay, which further shows a positive effect of PAC and CAC controlled follower quadcopter that minimizes the errors in the payload and its cable attitude. Thus, a stable human controlled flight is achieved.
\vspace{-0.05in}
\subsection{Simulation}
\vspace{-0.1in}
\begin{figure}[!htb]
    \centering
    \includegraphics[scale=0.95]{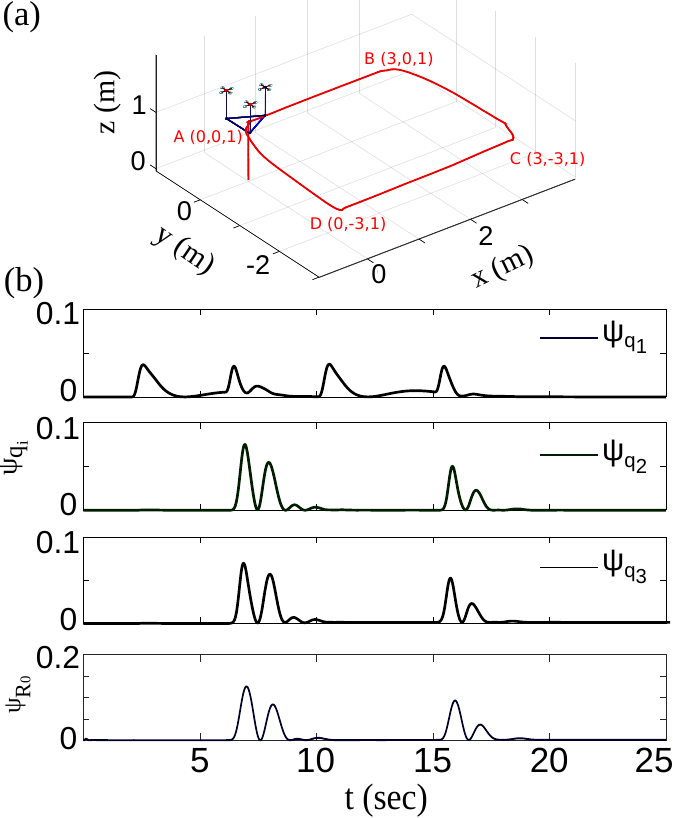}
    \caption{\small Simulation results for three quadcopters with cable-suspended payload system. Plot (a) shows the trajectory of the payload. The plots (b) and (c) show the cable and payload configuration error function for the system.\normalsize}
    \label{fig:simResults}
    \vspace{-4pt}
\end{figure}
In the experiment, a slender-rod shaped payload suspended at the end points was used. Since the points of suspension lie on a line, the roll attitude of the payload need not be controlled. However, for payloads not shaped like a slender rod, all three components of the payload's attitude need to be controlled along with the cable attitude control for followers. Thus, simulations for transportation of cable-suspended triangular shaped payload along a square trajectory, similar to the task in the experiment, using three quadcopters were performed. In particular, a $60 \, cm$ equilateral triangular shaped payload, mass $23 \, g$, is considered for the simulation. The leader quadcopter is commanded such that payload moves starting from point $O(0,0,0) \, m$ to points $A,B,C$, and $D$ as depicted in the Fig. \ref{fig:simResults} (a). A point to point position tracking controller \cite{pdposition} is used for the leader quadcopter. 

As shown in Fig. \ref{fig:simResults} (b), the peaks in the configuration error function for the cables and the payload attitudes, $\psi_{q_i}$ and $\psi_{R_0}$, appear when the leader quadcopter is commanded to change direction, i.e. at the vertices of the square trajectory. The rapid decay in the configuration error function depict the effectiveness of the CAC and PAC in minimizing the oscillations of the cables and the payload respectively.
\vspace{-0.05in}
\subsection{Discussion}
\vspace{-0.05in}
The main focus of this work was to implement a control architecture to enable human control of a multi-agent system. The results of the experiment demonstrate that, for the case of multiple quadcopters and a cable-suspended payload, it can be achieved by controlling the cables' and payload's attitude. Notably, the use of CAC and PAC will be tested further with multi-follower quadcopters in future work. Further, a state feedback controller is used in this study but there exists the possibility of other control designs which could enable better performance. The incorporation of human in the control loop can facilitate the navigation of quadcopters-payload system through unknown environments with potential applications in agriculture, warehouses, and tasks requiring impromptu payload transport. Therefore, the future work will focus on achieving the cables' and payload's attitude computation using portable onboard sensors, such as potentiometers, encoders, IMUs, etc., to eliminate the need of an indoor motion capture system and the ground station.
\vspace{-0.05in}
\section{CONCLUSIONS}
\vspace{-0.05in}
In this paper, we addressed the problem of manual control of a multiple quadcopter system collaboratively transporting a cable-suspended payload. As a human cannot control multiple quadcopters simultaneously, a leader-follower control architecture was developed, where a human directly controls the leader and followers are controlled by proposed Payload Attitude Controller (PAC) and Cable Attitude Controller (CAC). Human experiments were conducted for two quadcopters with a cable-suspended payload and simulations were performed using three quadcopters-payload system. The results showed successful human control of these systems to transport the payload spatially. With further testing of the control architecture and use of onboard sensors in future, the presented work demonstrates the possibility of incorporating human control of a multiple quadcopters and payload system to enable impromptu payload transportation in unknown environments.
\bibliographystyle{unsrt}
\bibliography{library.bib}
\end{document}